\documentclass[conference]{IEEEtran}
\IEEEoverridecommandlockouts
\usepackage{cite}
\usepackage{amsmath,amssymb,amsfonts}
\usepackage{algorithmic}
\usepackage{textcomp}
\usepackage[table,xcdraw,dvipsnames]{xcolor}
\definecolor{lightpink1}{HTML}{FFD5D9}
\definecolor{lightpink2}{HTML}{FFE2E6}
\definecolor{lightblue1}{HTML}{D6ECFF}  
\definecolor{lightblue2}{HTML}{E9F4FF}  

\definecolor{lavender1}{HTML}{DCC7E1} 
\definecolor{lavender2}{HTML}{BBA1CB} 
\definecolor{mauve1}{HTML}{A67EB7}   
\definecolor{mauve2}{HTML}{9B8EA9}  
\definecolor{violet1}{HTML}{7D5284}    
\definecolor{violet2}{HTML}{815C94}   
\definecolor{blueviolet1}{HTML}{757CBB}
\definecolor{blueviolet2}{HTML}{615EA8}
\definecolor{indigo1}{HTML}{4A4B9D}   
\definecolor{darkpurple1}{HTML}{422256}
\usepackage{url}
\usepackage{stfloats}
\usepackage{float} 
\usepackage[hidelinks, colorlinks=true, linkcolor=blue, citecolor=blue, breaklinks=true]{hyperref}
\usepackage[T1]{fontenc}
\usepackage{graphicx}
\usepackage{multirow}
\usepackage{makecell}
\usepackage{colortbl}
\usepackage{booktabs}
\usepackage[normalem]{ulem}
\usepackage{pifont}
\usepackage{dsfont}
\usepackage{subfigure}
\usepackage{subcaption}
\usepackage{stfloats}
\usepackage{marvosym}
\usepackage{amssymb} 
\usepackage{xcolor}
\usepackage[utf8]{inputenc}
\usepackage{adjustbox}
\usepackage{xcolor}
\usepackage{graphicx}

\newcommand{\legendbox}[1]{\raisebox{0.15ex}{\textcolor{#1}{\rule{0.6em}{0.6em}}}}

\definecolor{spleenc}{RGB}{255,0,0}
\definecolor{rkidneyc}{RGB}{0,180,0}
\definecolor{lkidneyc}{RGB}{255,215,0}
\definecolor{gallbladderc}{RGB}{0,102,255}
\definecolor{esophagusc}{RGB}{255,140,0}
\definecolor{liverc}{RGB}{128,0,200}
\definecolor{stomachc}{RGB}{0,220,220}
\definecolor{aortac}{RGB}{255,0,255}
\definecolor{ivcc}{RGB}{180,255,0}
\definecolor{psvc}{RGB}{255,170,200}
\definecolor{pancreasc}{RGB}{0,170,170}
\definecolor{ragc}{RGB}{200,160,255}
\definecolor{lagc}{RGB}{150,90,40}

\def\BibTeX{{\rm B\kern-.05em{\sc i\kern-.025em b}\kern-.08em
    T\kern-.1667em\lower.7ex\hbox{E}\kern-.125emX}}

\newcommand{\gain}[1]{
  \textcolor{gray}{\raisebox{0.15ex}{\scriptsize$\blacktriangle$}}\,#1
}
\newcommand{\downgain}[1]{
  \textcolor{gray}{\raisebox{0.05ex}{\scriptsize$\blacktriangledown$}}\,#1
}

\title{VCDP: Variation-Conditioned Distributional Proxy Learning for Semi-Supervised Medical Image Segmentation
}

\author{
\IEEEauthorblockN{
Zimu Zhang\textsuperscript{2},
Yiheng Zhong\textsuperscript{1},
Zhuoru Zhang\textsuperscript{2},\\
Yingzhen Hu\textsuperscript{2},
Yanan He\textsuperscript{1},
Fanliang Meng\textsuperscript{1},
Xiaofeng Liu\textsuperscript{1,\dag}
}

\IEEEauthorblockA{
\textsuperscript{1}Yale University, United States\\
\textsuperscript{2}Xi'an Jiaotong-Liverpool University, China\\
Email: xiaofeng.liu@yale.edu
}

\thanks{\textsuperscript{\dag}Corresponding author.}
}

\begin{document}
\maketitle

\begin{abstract}
Semi-supervised 3D medical image segmentation reduces the need for dense voxel-level annotations by exploiting unlabeled volumes. Although existing methods such as consistency regularization, pseudo-labeling, and co-training improve prediction-level robustness, they often provide insufficient feature-space organization for anatomically complex structures, especially small organs and ambiguous boundary regions with large intra-class variations. To address this issue, we propose Variation-Conditioned Distributional Proxy Learning (VCDP), a plug-and-play training-only regularization module for semi-supervised 3D medical image segmentation. VCDP represents each class with a learnable Gaussian distribution for shared class semantics and multiple variation prototypes for fine-grained intra-class patterns. A unified variation-conditioned compatibility score is further formulated to fuse distributional similarity and soft variation aggregation, guiding voxel embeddings to align with both global organ identity and local anatomical variations. VCDP is attached to decoder features during training and removed during inference, introducing no additional inference cost. Experiments on multi-organ segmentation benchmarks show that VCDP improves most evaluated baselines, particularly for small, ambiguous, and highly variable organs. Our anonymous code is released at \url{https://anonymous.4open.science/r/VCDP_code-41ED}.

\end{abstract}


\section{Introduction}
\label{sec:intro}

Medical image segmentation is a fundamental task in medical image analysis and plays an important role in clinical applications such as disease diagnosis, treatment planning, preoperative assessment, and therapy monitoring\cite{zhang2022pixelseg, qi2023mdf}. In 3D multi-organ segmentation, models are required to delineate multiple anatomical structures from volumetric CT or MRI scans, where accurate voxel-wise predictions are essential for reliable clinical decision-making. Despite the remarkable progress of deep learning-based segmentation methods, their success typically relies on dense voxel-level annotations, which are expensive and time-consuming to acquire and require substantial expertise from trained clinicians\cite{meng2022uncertainty, wu2023gcl, zhong2025pg}.

\begin{figure}[t]
    \centering
    \includegraphics[width=1\linewidth]{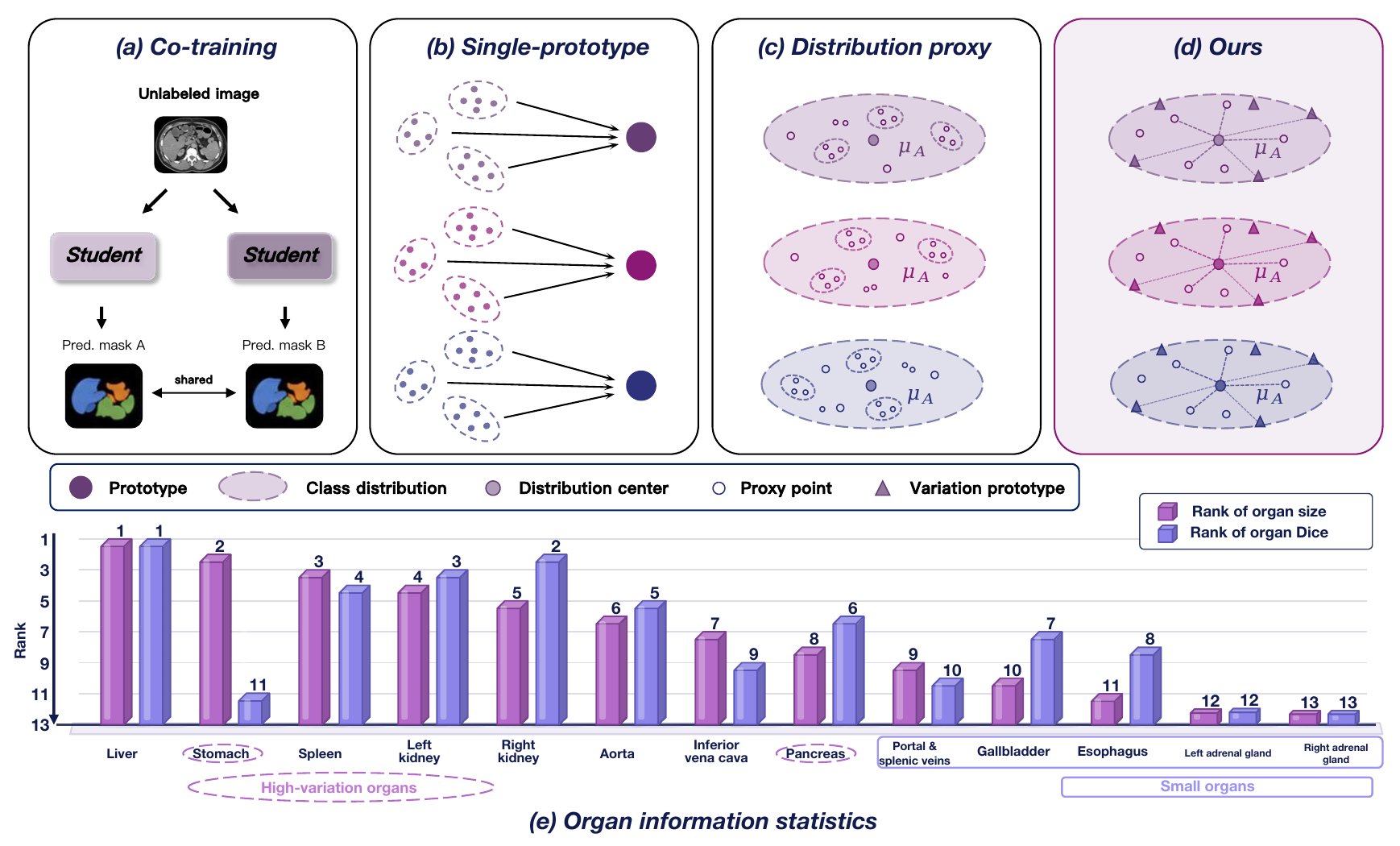}
    \caption{
    Motivation of the proposed VCDP framework. 
    (a) Co-training methods mainly enforce prediction-level consistency on unlabeled images. 
    (b) Single-prototype methods compress each class into one deterministic representative. 
    (c) Distribution proxy methods model class-level feature distributions but do not explicitly capture intra-class variation. 
    (d) Our method couples distributional proxies with intra-class variation prototypes to better represent fine-grained anatomical patterns. 
    (e) Organ-wise rankings in the Synapse dataset are shown in terms of both foreground organ size and segmentation Dice.
    }
    \label{fig:1}
\end{figure}

To mitigate the limitations caused by insufficient annotations, semi-supervised medical image segmentation has received increasing attention in recent years. It aims to reduce annotation cost by learning from a small set of labeled volumes together with abundant unlabeled data\cite{chen2022semi, duan2022rda, wang2023eye, zhang2022semi}. Representative methods include consistency regularization\cite{miyato2018virtual, sajjadi2016regularization, chi2024adaptivebidirectionaldisplacementsemisupervised}, pseudo-labeling\cite{sun2024fedmlpfederatedmultilabelmedical}, and co-training\cite{bai2023bidirectionalcopypastesemisupervisedmedical,sohn2020fixmatch}, which encourage models to learn from unlabeled images by enforcing stable or mutually agreed predictions. As illustrated in Fig.~\ref{fig:1}(a), a typical co-training framework feeds the same unlabeled image into two student networks and encourages their predicted masks to be consistent with each other. In this way, unlabeled data provides additional supervision through cross-model agreement, making the model less dependent on dense voxel-level annotations. Despite promising results, most of them primarily improve prediction-level robustness by enforcing consistency in the output space, while leaving the feature space insufficiently organized with respect to the complex anatomical variations within each semantic class.

This limitation is particularly pronounced for small organs and ambiguous boundary regions. Anatomical structures from the same class often exhibit strong intra-class variability across patients, slice locations, scanning conditions, and surrounding anatomical contexts. Their appearances can differ substantially in shape, scale, intensity, local texture, and boundary clarity\cite{yu2019uncertainty, li2020shape}. Moreover, small or elongated structures usually occupy only a limited number of voxels and are easily confused with adjacent tissues\cite{luo2021semi, assefa2025dycon, kumari2025annotation}. As a result, models may suffer from inter-class confusion, unstable boundary predictions, and insufficient recognition of small structures. As shown in Fig.~\ref{fig:1}(e), organs in Synapse\cite{landman2015miccai} exhibit clear disparities in both anatomical size and segmentation difficulty. Several small or high-variation organs, such as the esophagus, pancreas, and adrenal glands, are associated with lower segmentation ranks, indicating that these structures are particularly vulnerable to representation collapse and boundary errors. Notably, the stomach also shows a discrepancy between its relatively large anatomical size and its segmentation performance, suggesting that a large organ volume does not necessarily alleviate the difficulty caused by complex shape variations and ambiguous boundaries\cite{kumari2025annotation, luo2021semi}. Therefore, multi-organ segmentation requires careful consideration of both organ size and anatomical complexity. This observation motivates us to move beyond prediction-level consistency and investigate feature-space modeling for anatomically variable organs.

To better organize the feature space, proxy- and prototype-based representation learning provides a natural solution by encouraging intra-class compactness and inter-class separability. However, conventional single-prototype designs, as shown in Fig.~\ref{fig:1}(b), represent each class with one deterministic center, which may over-compress diverse anatomical patterns into an averaged representation\cite{kim2020proxyanchorlossdeep}. This is insufficient for organs with heterogeneous shapes, textures, and boundary appearances. Recent distributional proxy methods further model each class as a learnable distribution to capture class-level uncertainty, as illustrated in Fig.~\ref{fig:1}(c)\cite{zhou2022rethinkingsemanticsegmentationprototype}. Nevertheless, such global mean-variance representations mainly describe the overall class distribution and still lack explicit mechanisms to encode fine-grained intra-class variations, such as organ-specific sub-modes, ambiguous boundary patterns, and local contextual differences. Since recent proxy-based segmentation studies have shown the importance of intra-class variation modeling for resolving ambiguous regions\cite{BAE2026131783}, we argue that an effective medical segmentation proxy should jointly preserve distributional class semantics and expose multiple variation-aware directions within each class.

To this end, we propose \textbf{Variation-Conditioned Distributional Proxy Learning} (VCDP), a training-only regularization module for semi-supervised 3D medical image segmentation. As illustrated in Fig.~\ref{fig:1}(d), VCDP models every anatomical class as a distributional proxy conditioned by multiple intra-class variation prototypes. The distributional proxy, parameterized by a learnable Gaussian distribution, captures shared class-level semantics, while class-specific learnable variation prototypes model fine-grained intra-class patterns in the embedding space. By integrating these two complementary representations into a unified compatibility score, VCDP provides proxy-guided soft assignment for voxel embeddings and encourages them to match both the global identity of an organ and its local anatomical variations. Meanwhile, labeled voxels provide reliable semantic calibration for class representatives, preventing proxy drift during training. In this way, VCDP enables unlabeled data to participate in feature-space organization without requiring manual annotations, while improving the representation of diverse anatomical appearances and ambiguous boundary regions. As a plug-and-play auxiliary module, VCDP can be attached to decoder features during training and removed entirely during inference, without modifying the base semi-supervised framework or introducing additional computational cost.

Our contributions are summarized as follows:
\begin{itemize}
    \item We propose a plug-and-play variation-conditioned distributional proxy learning module for semi-supervised 3D medical image segmentation, which represents each anatomical class with a stochastic Gaussian proxy and multiple learnable variation prototypes to explicitly model class-level semantics and fine-grained anatomical variability in the feature space.

    \item We formulate a unified variation-conditioned class compatibility score that fuses distributional proxy similarity with soft variation aggregation, producing class-wise soft assignments that guide all voxel embeddings toward both global anatomical semantics and fine-grained intra-class patterns.

    \item We design a training-only proxy-guided regularization strategy that uses variation-conditioned soft assignments to organize all voxel embeddings, improving representation learning under limited annotations without adding inference cost.
    
\end{itemize}

\begin{figure*}[t]
    \centering
    \includegraphics[width=1\textwidth]{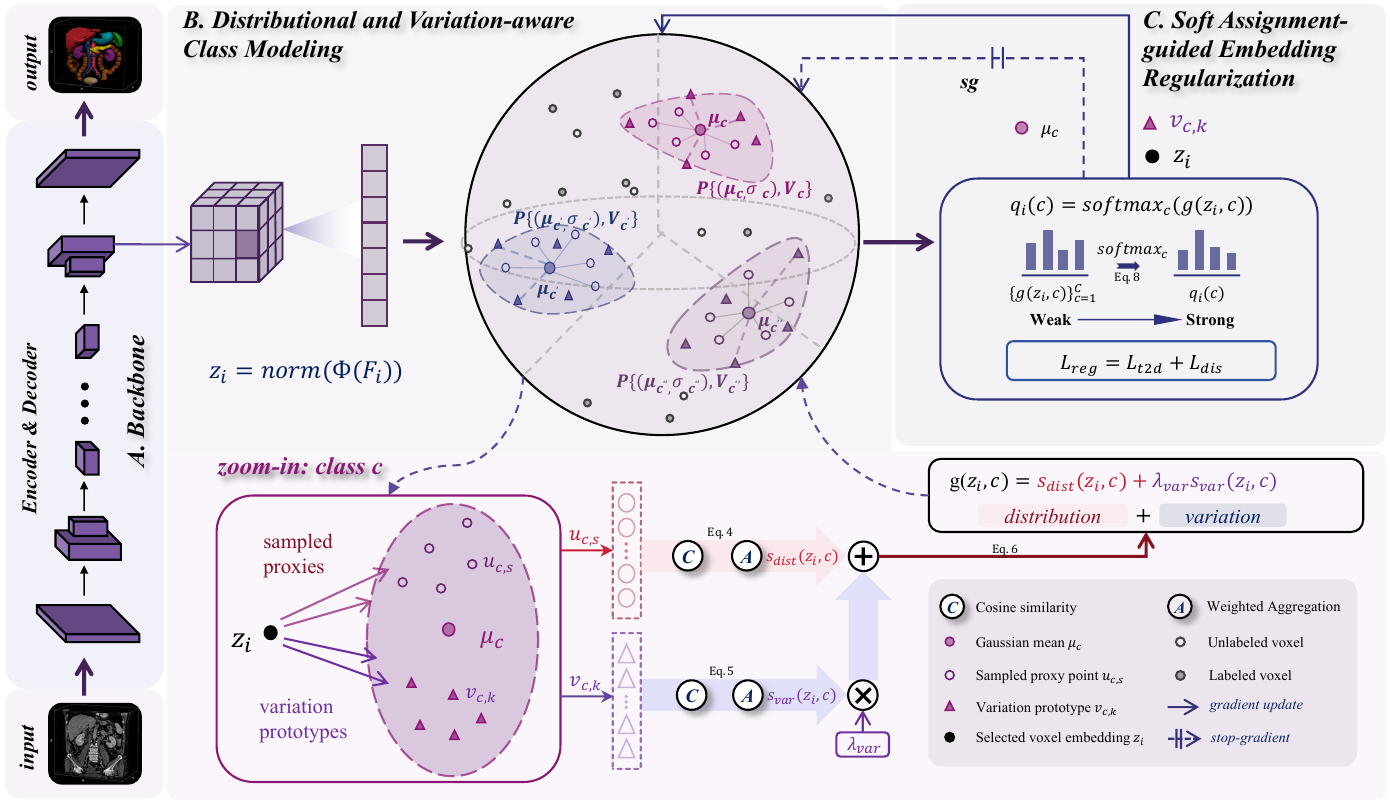}  
    \caption{Overview of the proposed VCDP framework for semi-supervised medical image segmentation.}
    \label{fig:2}
\end{figure*}

\section{Related Work}

\subsection{Semi-supervised Medical Image Segmentation}

Semi-supervised medical image segmentation reduces the reliance on dense annotations by exploiting unlabeled volumes together with limited labeled data. Existing methods mainly include consistency regularization, teacher-student learning, pseudo-labeling, uncertainty estimation, and co-training~\cite{yu2019uncertainty, wu2022exploringsmoothnessclassseparationsemisupervised, tarvainen2017mean, chen2021semi, chen2023magicnet, wang2023towards, wang2023dhc}. These methods improve unlabeled learning by enforcing stable predictions, generating pseudo labels, filtering uncertain supervision, or performing mutual supervision between networks. Despite their effectiveness, most of them focus on prediction-level constraints, leaving explicit feature-space representation learning relatively underexplored.

\subsection{Prototype and Proxy-based Representation Learning}

Prototype and proxy-based representation learning introduce class-level representatives to structure the embedding space and have been widely studied in metric learning, classification, and dense prediction~\cite{teh2020proxyncarevisitingrevitalizingproxy, kim2020proxyanchorlossdeep}. In semantic segmentation, prototypes regularize pixel- or voxel-level embeddings by encouraging intra-class compactness and inter-class separability~\cite{nunes2022segcontrast, petit2025disco}. Recent methods further enhance representation capacity with multiple prototypes, prototype refinement, or class-wise memory mechanisms~\cite{yang2024multimodalprototypesopenworldsemantic}, demonstrating the value of feature-space modeling for dense prediction.

\subsection{Distributional \& Variation-aware Representation Learning}

Distributional modeling captures uncertainty beyond deterministic feature points and has been explored for dense prediction with diverse appearances or ambiguous annotations. In medical image analysis, probabilistic representations have been used for uncertainty estimation, ambiguous segmentation, and prototype-based learning~\cite{BAE2026131783, kumari2025annotation}. Variation-aware modeling further captures fine-grained intra-class differences, such as local appearance changes, boundary ambiguity, and structural sub-modes. These representations are particularly useful for 3D medical segmentation, where voxel appearances vary across patients, spatial locations, and imaging conditions.

\section{Methodology}

\subsection{Overview}
Figure~\ref{fig:2} illustrates the overall framework of the proposed VCDP for semi-supervised 3D medical image segmentation. Unlike methods that mainly rely on prediction-level consistency or pseudo-label supervision, VCDP formulates SSMIS as a variation-conditioned anatomical representation learning problem, where each class is represented by a Gaussian proxy for shared semantics and multiple variation prototypes for intra-class structural sub-modes.

Let $\mathcal{X}=\{\mathcal{X}_l,\mathcal{X}_u\}$ denote a dataset of 3D medical images, where $\mathcal{X}_l=\{(x_i,y_i)\}_{i=1}^{N_l}$ is the labeled set and $\mathcal{X}_u=\{x_j\}_{j=N_l+1}^{N_l+N_u}$ is the unlabeled set, with $N_l \ll N_u$. Here, $x_i \in \mathbb{R}^{D \times H \times W}$ and $y_i \in \{0,1,\dots,C-1\}^{D \times H \times W}$ denote a 3D image and its voxel-wise label, respectively. Given an input image $x$, the segmentation network $f_{\theta}$ produces the prediction $p$ and an intermediate decoder feature map $F \in \mathbb{R}^{B \times C_f \times D \times H \times W}$. A lightweight projection head $\phi(\cdot)$ maps $F$ into normalized voxel embeddings $Z=\{z_i\}_{i=1}^{N}$, where $z_i \in \mathbb{R}^{d}$. Specifically, VCDP represents each class with a Gaussian proxy and multiple variation prototypes, as detailed in Section~\ref{B}, and uses their fused compatibility scores to regularize dense voxel embeddings from both labeled and unlabeled data, as described in Section~\ref{C}. It is jointly optimized with the original segmentation objective during training, while all auxiliary components are removed during inference with no additional cost.

\subsection{Distributional and Variation-aware Class Modeling}
\label{B}

Medical images often exhibit substantial anatomical ambiguity and inter-subject variability, making it insufficient to represent each anatomical class with a single deterministic prototype. Voxels belonging to the same class may present diverse appearances due to differences in organ shape, boundary uncertainty, lesion extent, and imaging contrast. Therefore, VCDP models each class with a variation-conditioned representation, where a Gaussian proxy captures shared class-level semantics under uncertainty and multiple variation prototypes describe fine-grained intra-class structural sub-modes.

For each anatomical class $c$, VCDP maintains a variation-conditioned class representation:
\begin{equation}
\mathcal{P}_c =
\{(\mu_c,\sigma_c), \mathbf{v}_{c,1}, \dots, \mathbf{v}_{c,K}\},
\end{equation}
where $K$ = 5 and $\mu_c$ denotes the learnable Gaussian proxy mean, $\sigma_c$ denotes a positive dispersion scale for stochastic proxy sampling, and $\{\mathbf{v}_{c,k}\}_{k=1}^{K}$ denotes learnable variation prototypes for intra-class structural sub-modes.

Given the voxel embeddings $z_i$, VCDP first constructs a distributional proxy for each anatomical class. Specifically, the Gaussian proxy of class $c$ is parameterized as
\begin{equation}
u_c \sim \mathcal{N}(\mu_c, \mathrm{diag}(\sigma_c^2)),
\end{equation}

where $\mu_c$ represents the learnable semantic center of the class, and $\sigma_c$ denotes a positive dispersion scale used to generate stochastic proxy representatives. In our implementation, $\sigma_c$ is kept fixed during training and is only used to control the sampling spread of the Gaussian proxy.

To estimate the distributional compatibility between a voxel embedding $z_i$ and class $c$, we draw $S$ stochastic class representatives from the Gaussian proxy:
\begin{equation}
u_{c,s} =
\mathrm{Normalize}(\mu_c + \sigma_c \odot \epsilon_s),
\quad
\epsilon_s \sim \mathcal{N}(0,I),
\quad s=1,\dots,S.
\end{equation}
The distributional similarity is then computed by averaging the cosine similarities between $z_i$ and all sampled representatives:
\begin{equation}
s_{\mathrm{dist}}(z_i,c)
=
\frac{1}{S}
\sum_{s=1}^{S}
\cos(z_i,u_{c,s}).
\end{equation}
Compared with a deterministic class prototype, the stochastic proxy provides a distributional description of class semantics and allows the model to account for uncertainty in voxel-level representations. This is particularly beneficial for semi-supervised 3D medical image segmentation, where unlabeled data may contain ambiguous boundaries and diverse anatomical variations.

In addition to the Gaussian proxy, each class is equipped with $K$ learnable variation prototypes to capture fine-grained intra-class structural patterns, such as local appearance changes, shape variations, and boundary ambiguity. For a voxel embedding $z_i$, VCDP computes its variation-aware similarity to class $c$ by softly aggregating the similarities over all class-specific variation prototypes:
\begin{equation}
s_{\mathrm{var}}(z_i,c)
=
\frac{1}{\tau}
\log
\sum_{k=1}^{K}
\exp\left(
\tau \cos(z_i,\mathbf{v}_{c,k})
\right),
\end{equation}

where $\tau$ controls the sharpness of the soft aggregation. This log-sum-exp formulation serves as a differentiable approximation to hard prototype selection, allowing each voxel embedding to adaptively match the most relevant intra-class variation patterns while still preserving gradient contributions from other possible structural modes.

After modeling distributional class semantics and intra-class variations, VCDP fuses them into a unified class compatibility score. For each voxel embedding $z_i$ and class $c$, the fused score is defined as
\begin{equation}
g(z_i,c)
=
s_{\mathrm{dist}}(z_i,c)
+
\lambda_{\mathrm{var}}
s_{\mathrm{var}}(z_i,c),
\end{equation}
where $\lambda_{\mathrm{var}}$ balances the contribution of the variation-aware term. This score jointly measures the compatibility between a voxel embedding and both the global semantic distribution and fine-grained structural variations of an anatomical class.

\begin{table*}[t!]
\centering
\small
\caption{
Quantitative comparison with state-of-the-art semi-supervised segmentation methods on Synapse (20\% labeled) and AMOS (5\% labeled). 
Light purple rows indicate methods equipped with VCDP. 
Gray values indicate performance changes relative to the corresponding baseline. 
Bold values denote better results within each baseline--Ours pair.
Large gains ($\geq 5$) are highlighted in red. 
DSC: Dice coefficient; NSD: normalized surface Dice; HD95: 95th percentile Hausdorff distance.
}
\label{tab1}
\setlength{\tabcolsep}{4pt}
\renewcommand{\arraystretch}{1}
\resizebox{0.85\textwidth}{!}{
\begin{tabular}{l|ccc|ccc}
\toprule
\multirow[c]{2}{*}{\textbf{Methods}}
& \multicolumn{3}{c|}{\cellcolor{mauve2}\textbf{Synapse (20\% labeled)}}
& \multicolumn{3}{c}{\cellcolor{mauve2}\textbf{AMOS (5\% labeled)}} \\

& \cellcolor{lavender1}\textbf{Dice}$\uparrow$
& \cellcolor{lavender2}\textbf{NSD}$\uparrow$
& \cellcolor{lavender1}\textbf{HD95}$\downarrow$
& \cellcolor{lavender1}\textbf{Dice}$\uparrow$
& \cellcolor{lavender2}\textbf{NSD}$\uparrow$
& \cellcolor{lavender1}\textbf{HD95}$\downarrow$ \\
\midrule

CPS~\cite{chen2021semi}        
& 66.26 & 75.48 & 13.41 & 57.95 & 66.76 & 14.13 \\

\cellcolor{mauve2!20}\textbf{CPS + Ours}
& \cellcolor{mauve2!20}\textbf{67.51}\gain{\textcolor{gray}{1.25}}
& \cellcolor{mauve2!20}\textbf{78.06}\gain{\textcolor{gray}{2.58}}
& \cellcolor{mauve2!20}\textbf{9.35} \downgain{\textcolor{gray}{4.06}}
& \cellcolor{mauve2!20}\textbf{59.16}\gain{\textcolor{gray}{1.21}}
& \cellcolor{mauve2!20}\textbf{68.24}\gain{\textcolor{gray}{1.48}}
& \cellcolor{mauve2!20}\textbf{13.25} \downgain{\textcolor{gray}{0.88}}\\

\midrule

MagicNet~\cite{chen2023magicnet}            
& 65.86 & 78.60 & \textbf{1.64} & 63.18 & 72.49 & 11.08 \\

\cellcolor{mauve2!20}\textbf{MagicNet + Ours}         
& \cellcolor{mauve2!20}\textbf{68.84}\gain{\textcolor{gray}{2.98}}
& \cellcolor{mauve2!20}\textbf{79.65} \gain{\textcolor{gray}{1.05}}
& \cellcolor{mauve2!20} 9.67 \gain{\textcolor{gray}{8.03}}
& \cellcolor{mauve2!20}\textbf{63.65}\gain{\textcolor{gray}{0.47}}
& \cellcolor{mauve2!20}\textbf{73.24}\gain{\textcolor{gray}{0.75}}
& \cellcolor{mauve2!20}\textbf{10.14}\downgain{\textcolor{gray}{0.94}} \\

\midrule

DHC~\cite{wang2023dhc}       
& 47.03 & 54.90 & 18.82 & \textbf{42.75} & 36.52 & 55.34 \\

\cellcolor{mauve2!20}\textbf{DHC + Ours}         
& \cellcolor{mauve2!20}\textbf{50.64}\gain{\textcolor{gray}{3.61}}
& \cellcolor{mauve2!20}\textbf{66.47}\gain{\textcolor{purple}{11.57}}
& \cellcolor{mauve2!20}\textbf{18.16}\downgain{\textcolor{gray}{0.66}}
& \cellcolor{mauve2!20} 42.63 \downgain{\textcolor{gray}{0.12}}
& \cellcolor{mauve2!20}\textbf{46.02}\gain{\textcolor{purple}{9.5}}
& \cellcolor{mauve2!20}\textbf{51.44} \downgain{\textcolor{gray}{3.9}}\\

\midrule

GenSSL~\cite{wang2023towards}      
& 57.28 & 70.73 & 7.69 & 42.81 & 33.32 & 51.14 \\

\cellcolor{mauve2!20}\textbf{GenSSL + Ours}         
& \cellcolor{mauve2!20}\textbf{63.74}\gain{\textcolor{purple}{6.46}}
& \cellcolor{mauve2!20}\textbf{74.93}\gain{\textcolor{gray}{2.2}}
& \cellcolor{mauve2!20}\textbf{5.86}\downgain{\textcolor{gray}{1.83}}
& \cellcolor{mauve2!20}\textbf{57.10}\gain{\textcolor{purple}{14.29}}
& \cellcolor{mauve2!20}\textbf{53.73}\gain{\textcolor{purple}{20.41}}
& \cellcolor{mauve2!20}\textbf{24.85} \downgain{\textcolor{purple}{26.29}}\\

\midrule

SS-Net~\cite{wu2022exploringsmoothnessclassseparationsemisupervised}           
& 49.22 & 55.62 & 40.31 & 30.87 & 33.45 & 69.97 \\

\cellcolor{mauve2!20}\textbf{SS-Net + Ours}
& \cellcolor{mauve2!20}\textbf{51.68}\gain{\textcolor{gray}{2.46}}
& \cellcolor{mauve2!20}\textbf{58.13}\gain{\textcolor{gray}{2.51}}
& \cellcolor{mauve2!20}\textbf{36.14}\downgain{\textcolor{gray}{4.17}}
& \cellcolor{mauve2!20}\textbf{51.26}\gain{\textcolor{purple}{20.39}}
& \cellcolor{mauve2!20}\textbf{55.84}\gain{\textcolor{purple}{22.39}}
& \cellcolor{mauve2!20}\textbf{41.72} \downgain{\textcolor{purple}{28.25}}\\

\midrule

Adsh~\cite{guo2022class}
& 32.48 & 40.06 & 53.19 & 38.22 & 41.36 & 60.15 \\

\cellcolor{mauve2!20}\textbf{Adsh + Ours}
& \cellcolor{mauve2!20}\textbf{39.44}\gain{\textcolor{purple}{6.96}}
& \cellcolor{mauve2!20}\textbf{48.28}\gain{\textcolor{purple}{8.23}}
& \cellcolor{mauve2!20}\textbf{45.63}\downgain{\textcolor{purple}{7.56}}
& \cellcolor{mauve2!20}\textbf{40.72}\gain{\textcolor{gray}{2.50}}
& \cellcolor{mauve2!20}\textbf{44.08}\gain{\textcolor{gray}{2.72}}
& \cellcolor{mauve2!20}\textbf{57.54}\downgain{\textcolor{gray}{2.61}} \\

\midrule

DCMamba~\cite{li2025diversityenhancedcollaborativemambasemisupervised}
& 45.82 & 49.15 & 17.70 & \textbf{48.91} & \textbf{55.18} & 22.01 \\

\cellcolor{mauve2!20}\textbf{DCMamba + Ours}
& \cellcolor{mauve2!20}\textbf{55.50}\gain{\textcolor{purple}{9.68}}
& \cellcolor{mauve2!20}\textbf{63.07}\gain{\textcolor{purple}{13.92}}
& \cellcolor{mauve2!20}\textbf{13.76}\downgain{\textcolor{gray}{3.94}} 
& \cellcolor{mauve2!20} 47.67 \downgain{\textcolor{gray}{1.24}}
& \cellcolor{mauve2!20} 52.64 \downgain{\textcolor{gray}{2.54}}
& \cellcolor{mauve2!20}\textbf{20.80}\downgain{\textcolor{gray}{1.21}} \\

\bottomrule
\end{tabular}
}
\end{table*}

\subsection{Soft Assignment-guided Embedding Regularization}
\label{C}

Given the projected embedding map \(Z\), we flatten its spatial dimensions into a set of voxel embeddings \(\{z_i\}_{i=1}^{N}\), where \(N=BD'H'W'\). 
VCDP regularizes these embeddings from both labeled and unlabeled volumes through class-wise soft assignments derived from the detached fused compatibility score.

To make the optimization stable, we use two complementary training paths: a dense embedding regularization path and a labeled proxy calibration path.

In the dense embedding regularization path, each voxel embedding is compared with every semantic class using the fused compatibility score. 
However, to prevent noisy unlabeled embeddings from directly shifting the class-level Gaussian proxy means, we stop the gradient of $\mu_c$ when computing this score:
\begin{equation}
\bar{g}(z_i,c)
=
g\left(
z_i,c;
\mathrm{sg}(\mu_c),
\sigma_c,
\{\mathbf{v}_{c,k}\}_{k=1}^{K}
\right).
\end{equation}

where \(\mathrm{sg}(\cdot)\) denotes the stop-gradient operation. 
Here, \(\bar{g}(z_i,c)\) is the stop-gradient version of the fused compatibility score used only in the dense regularization path. 
The distributional term \(s_{\mathrm{dist}}\) still measures the compatibility between \(z_i\) and the stochastic Gaussian representatives of class \(c\), but its Gaussian parameters \((\mu_c,\sigma_c)\) are not updated by this path. 
In contrast, the variation prototypes \(\{\mathbf{v}_{c,k}\}_{k=1}^{K}\) and the projected voxel embeddings remain trainable. 
Therefore, this path mainly shapes the embedding space and refines the intra-class variation prototypes, while preserving stable class-level Gaussian semantics.

Based on \(\bar{g}(z_i,c)\), we compute a class-wise soft assignment over all semantic classes:
\begin{equation}
q_i(c)
=
\frac{
\exp(\bar{g}(z_i,c))
}{
\sum_{c'=0}^{C-1}
\exp(\bar{g}(z_i,c'))
}.
\end{equation}
This assignment reflects the relative compatibility between voxel embedding \(z_i\) and each variation-conditioned class representation. 
Unlike confidence-based pseudo-labeling strategies that only supervise selected voxels with hard labels, this formulation provides dense feature-space regularization over all projected voxel embeddings in the current mini-batch.

The first regularization term encourages each voxel embedding to align with compatible class representations according to its soft assignment:
\begin{equation}
\mathcal{L}_{align}
=
\frac{1}{N}
\sum_{i=1}^{N}
\sum_{c=0}^{C-1}
q_i(c)
\left[
1-\bar{g}(z_i,c)
\right].
\end{equation}
This term promotes distribution-aware feature alignment by pulling each embedding toward class representations with high compatibility scores. 
To further enhance class discriminability, we introduce a distribution discrimination term:
\begin{equation}
m_c
=
\frac{1}{N}
\sum_{i=1}^{N}
\left(2q_i(c)-1\right)\bar{g}(z_i,c),
\end{equation}
\begin{equation}
\mathcal{L}_{dis}
=
\frac{1}{C}
\sum_{c=0}^{C-1}
\exp(-m_c).
\end{equation}
This term encourages embeddings strongly assigned to class \(c\) to have high compatibility with that class, while suppressing excessive responses from weakly assigned embeddings. 
The dense embedding regularization objective is then defined as
\begin{equation}
\mathcal{L}_{reg}
=
\mathcal{L}_{align}
+
\mathcal{L}_{dis}.
\end{equation}

In parallel, the labeled proxy calibration path provides a lightweight stabilization signal for the Gaussian proxy means by aligning them with reliable labeled voxel features.

For class \(c\), let
\begin{equation}
\Omega_c=\{i\mid y_i=c\}
\end{equation}
denotes the set of labeled voxel indices belonging to class \(c\) in the current mini-batch. 
We compute the labeled class feature anchor as
\begin{equation}
a_c
=
\mathrm{Normalize}
\left(
\frac{1}{|\Omega_c|}
\sum_{i\in\Omega_c}
z_i
\right).
\end{equation}
The proxy calibration loss is formulated as
\begin{equation}
\mathcal{L}_{cal}
=
\frac{1}{|\mathcal{C}_L|}
\sum_{c\in\mathcal{C}_L}
\left[
1-
\cos
\left(
\mathrm{Normalize}(\mu_c),
\mathrm{sg}(a_c)
\right)
\right],
\end{equation}
where \(\mathcal{C}_L\) denotes the set of classes present in the labeled mini-batch. 
The labeled anchor \(a_c\) is detached and serves as a reliable feature anchor, so this calibration path stabilizes the Gaussian class centers without directly imposing additional gradients on the backbone embeddings.

The final training objective is
\begin{equation}
\mathcal{L}
=
\mathcal{L}_{base}
+
\lambda_{reg}\mathcal{L}_{reg}
+
\lambda_{cal}\mathcal{L}_{cal},
\end{equation}
where \(\mathcal{L}_{base}\) denotes the original objective of the underlying semi-supervised segmentation framework, and \(\lambda_{reg}\) and \(\lambda_{cal}\) balance the two auxiliary objectives. 
During inference, the projection head, Gaussian proxies, and variation prototypes are removed, and the final prediction is produced only by the original segmentation network.

\section{Experiment} \label{sec:exp}

\subsection{Experimental Setup}

\noindent\textbf{Dataset.}
We evaluate VCDP on two multi-organ CT segmentation benchmarks, Synapse~\cite{landman2015miccai} and AMOS~\cite{ji2022amos}. Synapse contains 30 abdominal CT scans with 13 foreground organs, including spleen (Sp), right/left kidneys (RK/LK), gallbladder (Ga), esophagus (Es), liver (Li), stomach (St), aorta (Ao), inferior vena cava (IVC), portal and splenic veins (PSV), pancreas (Pa), and right/left adrenal glands (RAG/LAG). Following previous semi-supervised settings, we split Synapse into 20 training, 4 validation, and 6 testing scans, and use 20\% of the training scans as labeled data. AMOS contains 360 CT scans with 15 foreground organs. In addition to most Synapse organs, AMOS includes duodenum (Du), bladder (Bl), and prostate/uterus (P/U), while excluding portal and splenic veins. We use 216 scans for training, 24 for validation, and 120 for testing, and adopt a more challenging 5\% labeled setting with the remaining training scans used as unlabeled data.

\noindent\textbf{Implementation Details.}
We implement VCDP as a training-time plug-in module and integrate it into CPS\cite{chen2021semi}, MagicNet\cite{chen2023magicnet}, DHC\cite{wang2023dhc}, GenSSL\cite{wang2023towards}, SS-Net\cite{wu2022exploringsmoothnessclassseparationsemisupervised}, Adsh\cite{guo2022class}, and DCMamba\cite{li2025diversityenhancedcollaborativemambasemisupervised}. For fair comparison, we keep the original prediction branches and training objectives of all baselines unchanged, and jointly optimize VCDP as an auxiliary feature-space regularizer during training. All experiments are implemented in PyTorch and conducted on an NVIDIA A40 GPU with a batch size of 4. The weight decay for the proposed proxy/prototype module is set to $1\times10^{-4}$. At inference time, the projection head, Gaussian proxies, and variation prototypes are removed, thus introducing no additional computational cost. We report DSC, NSD, and HD95, where higher DSC/NSD and lower HD95 indicate better performance.

\begin{table*}[!t]
\centering
\tiny
\caption{
Per-class quantitative comparison with state-of-the-art semi-supervised segmentation methods on the Synapse multi-organ CT dataset.
Light purple rows indicate methods equipped with VCDP.
Dice scores are reported for 13 foreground organs.
Mean Dice denotes the average Dice coefficient over all foreground classes, and HD95 denotes the 95th percentile Hausdorff distance. Best organ-wise results are highlighted in red.
}
\label{tab:synapse_per_class}
\setlength{\tabcolsep}{3.0pt}
\renewcommand{\arraystretch}{1.20}
\resizebox{1.0\textwidth}{!}{
\begin{tabular}{l|ccccccccccccc|cc}
\toprule
\rowcolor{mauve2}
\textbf{Method}
& \textbf{Sp}
& \textbf{RK}
& \textbf{LK}
& \textbf{Ga}
& \textbf{Es}
& \textbf{Li}
& \textbf{St}
& \textbf{Ao}
& \textbf{IVC}
& \textbf{PSV}
& \textbf{Pa}
& \textbf{RAG}
& \textbf{LAG}
& \textbf{Dice[\%]}$\uparrow$
& \textbf{HD95[mm]}$\downarrow$ \\
\midrule

CPS~\cite{chen2021semi}
& 84.19 & 91.26 & 89.13 & 22.78 & 40.45 & 92.79
& 65.23 & 79.11 & 80.81 & 65.23 & 47.43 & 46.00 & 57.01
& 66.26 & 13.41 \\

\rowcolor{mauve2!20}
\textbf{CPS + Ours}
& 85.80 & 90.25 & 90.79 & 29.63 & 45.92 & \textcolor{red}{\textbf{93.19}} & 62.65 & 80.00 & 82.13 & 63.80 & 53.28 & 49.15 & 51.01
& 67.51 & 9.35 \\

\midrule

MagicNet~\cite{chen2023magicnet}
& 81.95 & 90.95 & 90.19 & 26.13 & 42.38 & 90.75 & 57.22 & 80.49 & 80.25 & 63.93 & 42.34 & 48.27 & 61.31
& 65.86 & 1.64 \\

\rowcolor{mauve2!20}
\textbf{MagicNet + Ours}
& 84.33 & \textcolor{red}{\textbf{91.40}} & \textcolor{red}{\textbf{90.86}} & 27.48 & 50.54 & 90.31 & \textcolor{red}{\textbf{65.50}} & 78.20 & \textcolor{red}{\textbf{83.66}} & \textcolor{red}{\textbf{68.35}} & 52.72 & 47.74 & \textcolor{red}{\textbf{63.80}}
& \textcolor{red}{\textbf{68.84}} & 9.67 \\

\midrule

DHC~\cite{wang2023dhc}
& 77.90 & 51.60 & 48.70 & 55.50 & 52.60 & 86.80
& 35.90 & 63.20 & 55.00 & 35.60 & 22.60 & 21.10 & 4.80
& 47.03 & 18.82 \\

\rowcolor{mauve2!20}
\textbf{DHC + Ours}
& 54.90 & 53.40 & 63.30 & 63.50 & 63.30 & 89.50
& 32.00 & 82.20 & 23.30 & 36.80 & 73.10 & 11.10 & 11.80
& 50.64 & 18.16 \\

\midrule

GenSSL~\cite{wang2023towards}
& 76.97 & 72.60 & 66.00 & 50.64 & 60.66 & 91.08 & 50.33 & 79.94 & 74.67 & 51.70 & 46.53 & 44.64 & 19.50
& 57.28 & 7.69 \\

\rowcolor{mauve2!20}
\textbf{GenSSL + Ours}
& 80.08 & 74.23 & 74.01 & 44.62 & 65.04 & 91.96 & 55.79 & \textcolor{red}{\textbf{83.02}} & 76.24 & 56.33 & 50.81 & \textcolor{red}{\textbf{51.19}} & 25.24
& 63.74 & \textcolor{red}{\textbf{5.86}} \\

\midrule

SS-Net~\cite{wu2022exploringsmoothnessclassseparationsemisupervised}
& 59.51 & 44.70 & 43.32 & 62.01 & 51.43 & 87.03 & 45.33 & 70.01 & 23.13 & 41.33 & 68.41 & 4.90 & 38.81
& 49.22 & 40.31 \\

\rowcolor{mauve2!20}
\textbf{SS-Net + Ours}
& 64.21 & 62.53 & 26.61 & \textcolor{red}{\textbf{65.11}} & \textcolor{red}{\textbf{67.13}} & 88.80 & 24.51 & 79.71 & 24.73 & 44.64 & \textcolor{red}{\textbf{83.71}} & 7.71 & 32.61
& 51.68 & 36.14 \\

\midrule

Adsh~\cite{guo2022class}
& 41.10 & 0.00 & 7.01 & 55.23 & 58.52 & 90.33 & 30.30 & 54.81 & 16.41 & 0.00 & 68.71 & 0.00 & 0.00
& 32.48 & 53.19 \\

\rowcolor{mauve2!20}
\textbf{Adsh + Ours}
& 39.60 & 0.00 & 51.81 & 43.14 & 58.43 & 89.01 & 27.73 & 69.83 & 25.00 & 36.63 & 71.61 & 0.00 & 0.00
& 39.44 & 45.63 \\

\midrule

DCMamba~\cite{li2025diversityenhancedcollaborativemambasemisupervised}
& 80.04 & 78.22 & 81.48 & 7.63 & 38.11 & 91.91 & 56.39 & 75.72 & 50.74 & 19.10 & 16.28 & 0.00 & 0.00
& 45.82 & 17.70 \\

\rowcolor{mauve2!20}
\textbf{DCMamba + Ours}
& \textcolor{red}{\textbf{86.49}} & 69.62 & 73.52 & 55.49 & 55.35 & 92.94 & 61.37 & 72.21 & 47.51 & 39.40 & 29.30 & 11.81 & 26.46
& 55.50 & 13.76 \\

\bottomrule
\end{tabular}
}
\vspace{1pt}

\end{table*}

\begin{table*}[!t]
\centering
\tiny
\caption{
Per-class quantitative comparison on the AMOS multi-organ CT dataset.
Light purple rows indicate methods equipped with VCDP.
Dice scores are reported for 15 foreground organs.
Mean Dice denotes the average Dice coefficient over all foreground classes, and HD95 denotes the 95th percentile Hausdorff distance. Best organ-wise results are highlighted in red.
}
\label{tab:amos_per_class}
\setlength{\tabcolsep}{2.4pt}
\renewcommand{\arraystretch}{1.18}
\resizebox{1.0\textwidth}{!}{
\begin{tabular}{l|ccccccccccccccc|cc}
\toprule
\rowcolor{mauve2}
\textbf{Method}
& \textbf{Sp}
& \textbf{RK}
& \textbf{LK}
& \textbf{Ga}
& \textbf{Es}
& \textbf{Li}
& \textbf{St}
& \textbf{Ao}
& \textbf{IVC}
& \textbf{Pa}
& \textbf{RAG}
& \textbf{LAG}
& \textbf{Du}
& \textbf{Bl}
& \textbf{P/U}
& \textbf{Dice[\%]}$\uparrow$
& \textbf{HD95[mm]}$\downarrow$ \\
\midrule

CPS~\cite{chen2021semi}
& 77.44 & 78.82 & 77.93 & 46.67 & 40.61 & 88.00 & 50.15 & 75.62 & 60.60 & 48.94 & 42.09 & 34.57 & 33.74 & 69.20 & 44.89
& 57.95 & 14.13 \\

\rowcolor{mauve2!20}
\textbf{CPS + Ours}
& 80.75 & 79.86 & 78.60 & 46.39 & 48.75 & 88.42 & 51.01 & 75.95 & 62.47 & 50.74 & 43.77 & 33.53 & 33.04 & 68.40 & 45.69
& 59.16 & 13.25 \\

\midrule

MagicNet~\cite{chen2023magicnet}
& 80.52 & 83.45 & 83.95 & 48.35 & 50.35 & 87.61 & 60.38 & \textcolor{red}{\textbf{84.32}} & \textcolor{red}{\textbf{72.62}} & 57.04 & 48.40 & 38.53 & 42.04 & 66.16 & 47.86
& 63.18 & 11.08 \\

\rowcolor{mauve2!20}
\textbf{MagicNet + Ours}
& 78.78 & 84.85 & 86.03 & \textcolor{red}{\textbf{49.78}} & \textcolor{red}{\textbf{54.19}} & 87.87 & 55.56 & 82.83 & 68.52 & 56.08 & \textcolor{red}{\textbf{49.33}} & \textcolor{red}{\textbf{42.95}} & 37.39 & 71.05 & \textcolor{red}{\textbf{49.46}}
& \textcolor{red}{\textbf{63.65}} & \textcolor{red}{\textbf{10.14}} \\

\midrule

DHC~\cite{wang2023dhc}
& 69.70 & 54.20 & 58.90 & 6.70 & 20.30 & 75.10 & 41.40 & 74.20 & 57.10 & 47.70 & 0.00 & 0.00 & 28.60 & 62.60 & 44.80
& 42.75 & 55.34 \\

\rowcolor{mauve2!20}
\textbf{DHC + Ours}
& 64.12 & 59.63 & 62.02 & 6.73 & 17.91 & 78.61 & 46.30 & 71.92 & 58.52 & 51.41 & 0.00 & 0.00 & 26.01 & 61.73 & 34.90
& 42.63 & 51.44 \\

\midrule

GenSSL~\cite{wang2023towards}
& 69.20 & 67.95 & 73.13 & 41.99 & 2.52 & 78.61 & 51.86 & 74.74 & 52.79 & 48.07 & 0.00 & 0.00 & 27.89 & 34.20 & 19.17
& 42.81 & 51.14 \\

\rowcolor{mauve2!20}
\textbf{GenSSL + Ours}
& 83.54 & 80.86 & 79.08 & 32.73 & 49.75 & 84.70 & 51.83 & 82.50 & 60.63 & \textcolor{red}{\textbf{57.14}} & 37.43 & 40.80 & 35.95 & 60.34 & 19.17
& 57.10 & 24.85 \\

\midrule

SS-Net~\cite{wu2022exploringsmoothnessclassseparationsemisupervised}
& 42.12 & 45.73 & 49.20 & 6.71 & 0.00 & 80.60 & 36.91 & 58.83 & 27.31 & 36.32 & 0.00 & 0.00 & 9.12 & 51.22 & 19.21
& 30.87 & 69.97 \\

\rowcolor{mauve2!20}
\textbf{SS-Net + Ours}
& \textcolor{red}{\textbf{88.21}} & \textcolor{red}{\textbf{85.53}} & \textcolor{red}{\textbf{89.01}} & 7.01 & 0.00 & \textcolor{red}{\textbf{90.20}} & \textcolor{red}{\textbf{57.83}} & 81.43 & 67.31 & 55.71 & 0.00 & 35.42 & \textcolor{red}{\textbf{45.60}} & 62.71 & 3.00
& 51.26 & 41.72 \\

\midrule

Adsh~\cite{guo2022class}
& 65.21 & 62.80 & 52.01 & 6.72 & 0.00 & 83.91 & 38.90 & 66.61 & 58.82 & 36.01 & 0.00 & 0.00 & 14.63 & 68.64 & 19.21
& 38.22 & 60.15 \\

\rowcolor{mauve2!20}
\textbf{Adsh + Ours}
& 68.81 & 67.82 & 66.80 & 6.73 & 0.00 & 84.23 & 40.74 & 71.40 & 55.81 & 39.71 & 0.00 & 0.00 & 18.50 & \textcolor{red}{\textbf{71.32}} & 19.21
& 40.72 & 57.54 \\

\midrule

DCMamba~\cite{li2025diversityenhancedcollaborativemambasemisupervised}
& 75.15 & 69.21 & 68.06 & 43.65 & 34.92 & 84.66 & 50.68 & 70.80 & 47.12 & 39.78 & 24.06 & 14.26 & 28.87 & 51.91 & 30.44
& 48.91 & 22.01 \\

\rowcolor{mauve2!20}
\textbf{DCMamba + Ours}
& 75.90 & 66.95 & 66.60 & 42.89 & 34.18 & 84.12 & 49.70 & 71.68 & 46.56 & 40.26 & 24.54 & 0.00 & 28.44 & 52.38 & 30.81
& 47.67 & 20.80 \\

\bottomrule
\end{tabular}
}
\vspace{1pt}

\end{table*}

\subsection{Comparisons with State-of-the-art Methods}

As shown in Table~\ref{tab1}, VCDP brings broad improvements across different semi-supervised segmentation frameworks on Synapse under the 20\% labeled setting. When integrated into CPS, VCDP improves Dice to 67.51\% $(\uparrow 1.25\%)$, increases NSD from 75.48\% to 78.06\% $(\uparrow 2.58\%)$, and reduces HD95 from 13.41 mm to 9.35 mm $(\downarrow 4.06\text{ mm})$. Similar gains are observed for MagicNet, where Dice improves to 68.84\% $(\uparrow 2.98\%)$. More substantial gains are observed for baselines with lower initial performance, with Dice improvements of 3.61\% for DHC, 6.46\% for GenSSL, 2.46\% for SS-Net, 6.96\% for Adsh, and 9.68\% for DCMamba. In particular, DCMamba+VCDP achieves a large Dice gain $(\uparrow 9.68\%)$, while GenSSL+VCDP reduces HD95 to 5.86 $(\downarrow 1.83\text{ mm})$. These results show that VCDP is not tied to a specific semi-supervised strategy, but provides complementary feature-space supervision for diverse prediction-space frameworks.

On AMOS under the more challenging 5\% labeled setting, VCDP also improves most evaluated baselines. For MagicNet, VCDP improves Dice to 63.65\% $(\uparrow 0.47\%)$ and reduces HD95 from 11.08 mm to 10.14 mm $(\downarrow 0.94\text{ mm})$, achieving the best overall performance among the evaluated methods. The gains are more promising for GenSSL and SS-Net: GenSSL+VCDP improves Dice/NSD by 14.29\%/20.41\% and reduces HD95 by 26.29 mm, while SS-Net+VCDP improves Dice/NSD by 20.39\%/22.39\%. These large improvements indicate that VCDP provides strong representation-level guidance when the baseline feature space is insufficiently organized. For DHC and DCMamba, Dice improvements are less uniform on AMOS, but VCDP still reduces HD95, suggesting potential benefits in boundary robustness.

Tables~\ref{tab:synapse_per_class} and~\ref{tab:amos_per_class} further report the per-class Dice scores on Synapse and AMOS, respectively. The gains are mainly concentrated on challenging organs with small sizes, ambiguous boundaries, and large intra-class variations. On Synapse, VCDP notably improves difficult structures, including the gallbladder, esophagus, stomach, and pancreas. On AMOS, VCDP brings substantial gains to under-performing baselines such as GenSSL and SS-Net. Overall, these results indicate that VCDP improves organ-level robustness, with particularly strong benefits for anatomically difficult categories requiring fine-grained variation modeling.

\begin{figure}[t]
    \centering
    \includegraphics[width=0.95\columnwidth]{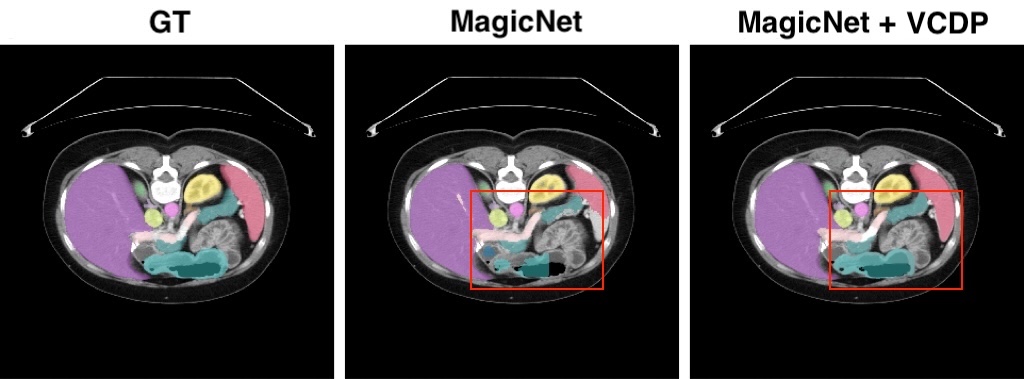}
    \caption{
    Qualitative comparison on the 20\% labeled Synapse dataset, with a focus on the pancreas.
    Red boxes highlight the pancreas region.
    Compared with MagicNet, VCDP produces a more complete pancreas mask that is closer to the ground truth.
    Organ legend:
    \legendbox{spleenc} spleen,
    \legendbox{rkidneyc} right kidney,
    \legendbox{lkidneyc} left kidney,
    \legendbox{gallbladderc} gallbladder,
    \legendbox{esophagusc} esophagus,
    \legendbox{liverc} liver,
    \legendbox{stomachc} stomach,
    \legendbox{aortac} aorta,
    \legendbox{ivcc} inferior vena cava,
    \legendbox{psvc} portal/splenic vein,
    \legendbox{pancreasc} pancreas,
    \legendbox{ragc} right adrenal gland, and
    \legendbox{lagc} left adrenal gland.
    }
    \label{fig:qualitative}
\end{figure}

\subsection{Qualitative Analysis}
Fig.~\ref{fig:qualitative} compares the segmentation results of the ground truth, MagicNet, and MagicNet+VCDP on the pancreas region. MagicNet produces fragmented and inaccurate predictions around the pancreas, reflecting its difficulty in handling small and ambiguous anatomical structures. In contrast, MagicNet+VCDP better preserves the pancreas region and generates a more complete mask closer to the ground truth. These results show that VCDP improves the segmentation of challenging organs through variation-conditioned feature-space modeling.

\subsection{Ablation Study}
To validate the effectiveness of Gaussian and variation modeling in VCDP, we conduct ablation studies on the Synapse dataset using MagicNet as the baseline framework, as shown in Table~\ref{tab:ablation}. The baseline MagicNet achieves 65.86\% mean Dice and 78.60\% NSD. With only the Gaussian proxy, the performance increases to 68.06\% Dice ($\uparrow$ 2.20\%) and 79.25\% NSD ($\uparrow$ 0.65\%), showing the benefit of distributional class modeling. With only variation prototypes, the model also improves the baseline Dice to 66.96\% ($\uparrow$ 1.10\%), indicating that intra-class variation modeling provides useful feature-space guidance. Combining both components achieves the best results, reaching 68.84\% Dice ($\uparrow$ 2.98\%) and 79.65\% NSD ($\uparrow$ 1.05\%). These results verify that Gaussian proxies and variation prototypes are complementary for modeling shared class semantics and fine-grained anatomical variations.

\begin{table}[t]
\centering
\small
\caption{
Ablation study on the effectiveness of Gaussian and variation modeling in VCDP on the Synapse dataset.
G and V denote the Gaussian proxy and variation prototypes, respectively.
}
\label{tab:ablation}
\begin{tabular}{c|cc|cc}
\toprule
 & G & V & Dice [\%]$\uparrow$ & NSD [\%]$\uparrow$ \\
\midrule
I   &   &   & 65.86 & 78.60 \\
II  & \checkmark &   & 68.06 & 79.25 \\
III &   & \checkmark & 66.96 & 78.60 \\
IV  & \checkmark & \checkmark & \textbf{68.84} & \textbf{79.65} \\
\bottomrule
\end{tabular}
\end{table}

\section{Conclusion}

In this paper, we proposed VCDP, a plug-and-play training-only regularization module for semi-supervised 3D medical image segmentation. By representing each anatomical class with a Gaussian proxy and multiple variation prototypes, VCDP explicitly organizes voxel embeddings to capture both shared class semantics and fine-grained intra-class variations. Experiments on Synapse and AMOS demonstrate broad improvements across multiple semi-supervised frameworks, especially for anatomically challenging organs. Since all auxiliary components are removed during inference, VCDP introduces no additional inference cost.

\bibliographystyle{IEEEtran}
\bibliography{main}
\end{document}